\documentclass[lettersize,journal]{IEEEtran}
\usepackage{amsmath,amsfonts}
\usepackage{listings}
\usepackage[ruled,vlined,linesnumbered]{algorithm2e}
\usepackage[dvipsnames]{xcolor}
\usepackage{array}
\usepackage[caption=false,font=normalsize,labelfont=sf,textfont=sf]{subfig}
\usepackage{textcomp}
\usepackage{stfloats}
\usepackage{url}
\usepackage{verbatim}
\usepackage{graphicx}
\usepackage{cite}
\usepackage{amssymb}
\usepackage{amsmath}
\usepackage{booktabs}
\usepackage{hyperref}
\usepackage{subfig}
\usepackage{subfloat}
\usepackage{enumerate}
\usepackage{threeparttable}
\usepackage{color}
\usepackage{makecell}
\usepackage{multirow}
\usepackage{bm}
\usepackage{bbding}

\begin{document}

\title{SiT-MLP: A Simple MLP with Point-wise Topology Feature Learning for Skeleton-based Action Recognition}

\author{Shaojie~Zhang,~Jianqin~Yin,~\IEEEmembership{Member,~IEEE},~Yonghao~Dang~and~Jiajun~Fu
\thanks{Corresponding author: Jianqin Yin}
\thanks{Shaojie Zhang, Jianqin Yin, Yonghao Dang and Jiajun Fu are with the School of Artificial Intelligence, Beijing University of Posts and Telecommunications, Beijing 100876, China (e-mail:zsj@bupt.edu.cn; jqyin@bupt.edu.cn;  dyh2018@bupt.edu.cn; JaakkoFu@bupt.edu.cn;)}}

\markboth{IEEE Transactions on Circuits and Systems for Video Technology}%
{Shell \MakeLowercase{\textit{et al.}}: Bare Demo of IEEEtran.cls for IEEE Journals}

\maketitle

\begin{abstract}
Graph convolution networks (GCNs) have achieved remarkable performance in skeleton-based action recognition. However, previous GCN-based methods rely on elaborate human priors excessively and construct complex feature aggregation mechanisms, which limits the generalizability and effectiveness of networks. To solve these problems, we propose a novel Spatial Topology Gating Unit (STGU), an MLP-based variant without extra priors, to capture the co-occurrence topology features that encode the spatial dependency across all joints. In STGU, to learn the point-wise topology features, a new gate-based feature interaction mechanism is introduced to activate the features point-to-point by the attention map generated from the input sample. Based on the STGU, we propose the first MLP-based model, SiT-MLP, for skeleton-based action recognition in this work. Compared with previous methods on three large-scale datasets, SiT-MLP achieves competitive performance. In addition, SiT-MLP reduces the parameters significantly with favorable results. The code will be available at https://github.com/BUPTSJZhang/SiT-MLP.
\end{abstract}


\section{Introduction}
\IEEEPARstart{H}{uman} action recognition is an essential task in computer vision. It can enhance robot intelligence and improve the communication efficiency of human-computer interaction \cite{ren2020survey}. In recent years, thanks to the development of depth sensors \cite{zhang2012microsoft} and human pose estimation algorithms \cite{cao2017realtime}, it is easy to capture the sequence of human skeletons. Due to the robustness of human skeletons to background clutter and illumination changes, skeleton-based action recognition has attracted much interest \cite{wang2023comprehensive}.

\begin{figure}
    \centering
    \includegraphics[scale=0.52]{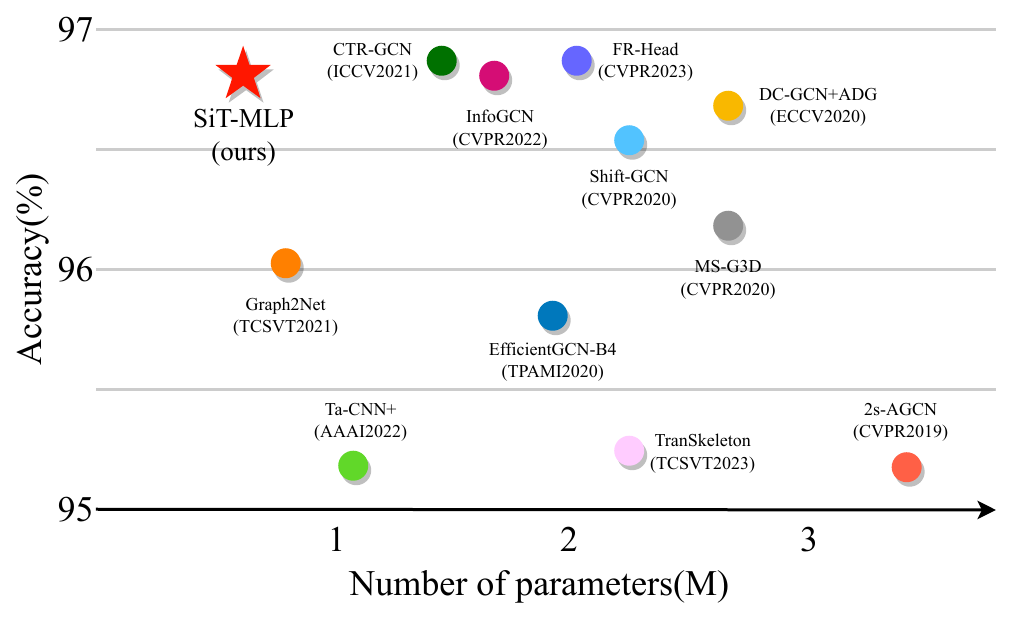}
    \caption{Comparsion of performance and parameter size on X-sub benchmark of NTU RGB+D 60 dataset. We report the accuracy as performance on the vertical dimension. The closer to the top-left, the better. Our method (SiT-MLP, in red) archives the highest performance with the fewest parameters.}
    \label{fig: R}
\end{figure}

Yan et al. \cite{yan2018spatial} first propose to build a human topology graph, treating joints and their explicit connections as nodes and edges respectively, and apply Graph Convolutional Network (GCN) \cite{kipf2016semi} on such a predefined graph to capture spatial-temporal co-occurrence features. The predefined graph introduces priors of human natural connections, and GCN can aggregate the neighboring joints' information and update the current joint's features. Since then, GCN-based methods \cite{yan2018spatial,cheng2020skeleton,cheng2020decoupling,liu2020disentangling,hao2021hypergraph,chen2021multi,zhang2020semantics,ye2020dynamic,shi2019two,chen2021channel,song2022constructing,chi2022infogcn,pan2022view} have become a research paradigm for skeleton-based action recognition. 

However, recent GCN-based approaches \cite{chen2021channel,chi2022infogcn,ye2020dynamic,liu2020disentangling,wu2021graph2net,xiong2022human,miao2021central} tend to construct complex feature aggregation methods to improve performance. As shown in Fig. \ref{fig: R}, the improved performance has also led to an increase in parameters. These approaches are always heavyweight and require additional elaborate priors. The vast number of parameters of the complex feature aggregations usually make the network not efficient. Moreover, the priors are related to the order of labeled joints and their physical connections. The introduction of the priors makes their network difficult to modify and generalize. Thus, a question naturally arises: “\textit{Can we tackle the skeleton-based action recognition without any priors and complex aggregations?}” 

\begin{figure}
    \centering
    \includegraphics[scale=0.275]{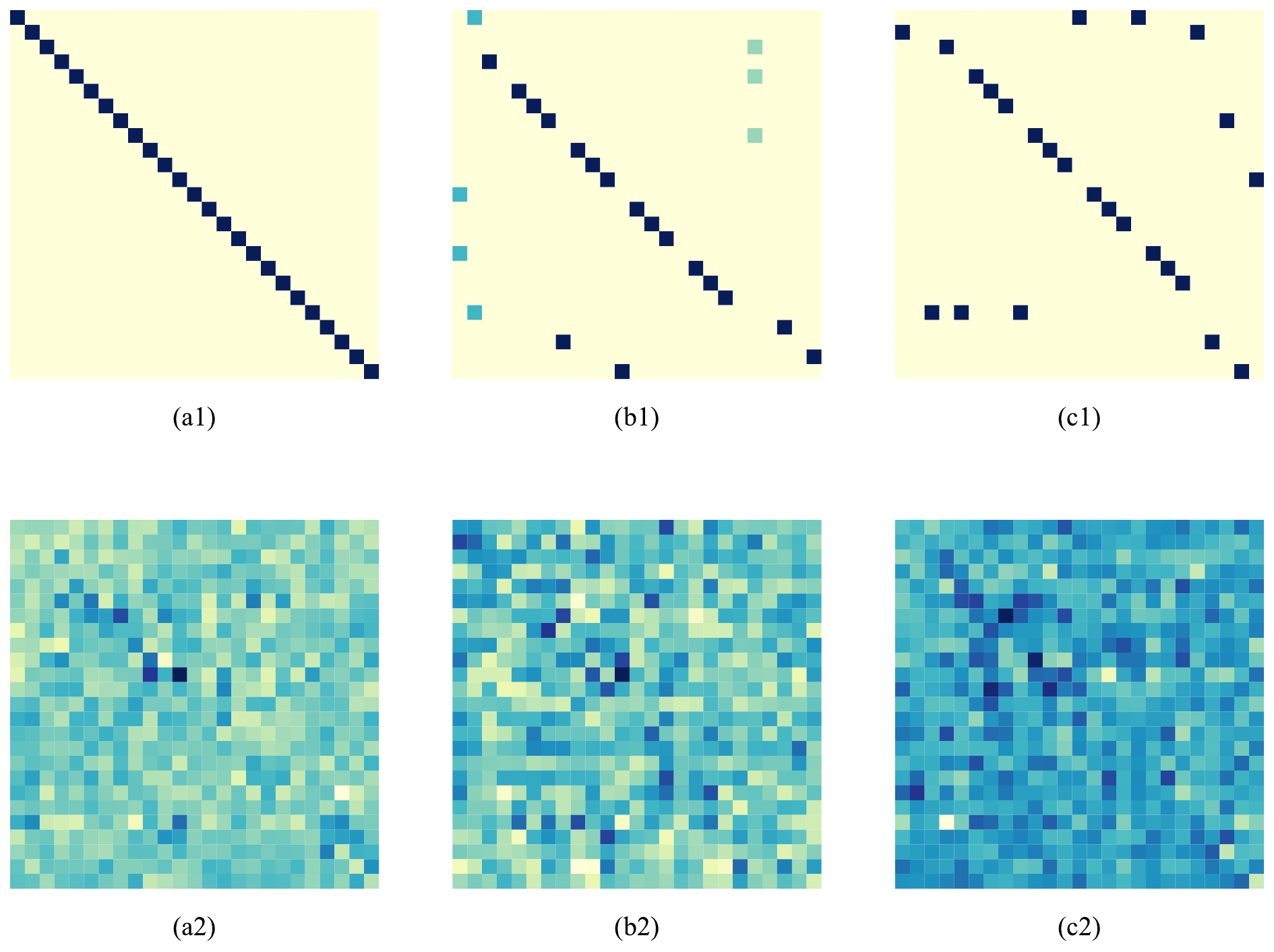}
    \caption{The comparison between the initialized normalized adjacency matrices and the final optimized adjacency matrices in the previous method \cite{chen2021channel}. The letters a,b, and c denote the self-link matrix, the inward connections, and the outward connections matrix, respectively. The numbers 1 and 2 indicate the initialized adjacency matrix and the final adjacency matrix, respectively.}
    \label{fig: A}
\end{figure}

To answer this question, we first review why GCN can achieve such success in skeleton-based action recognition. In most previous methods, the topology connections are set as learnable to break the limitations of the static topology attention and model the implicit connections. As shown in Fig. \ref{fig: A}, the final optimized matrices in \cite{chen2021channel} tend to be the global relationship between any two joints. In other words, the reason for the favorable performance of GCN-based approaches may not be the priors but the modeling of the relationships between any two joints. Thus, we think other structures, for example, \textbf{the simple MLP can also model such global relationships}.

\begin{figure*} [htbp]
    \centering
    \includegraphics[scale=0.52]{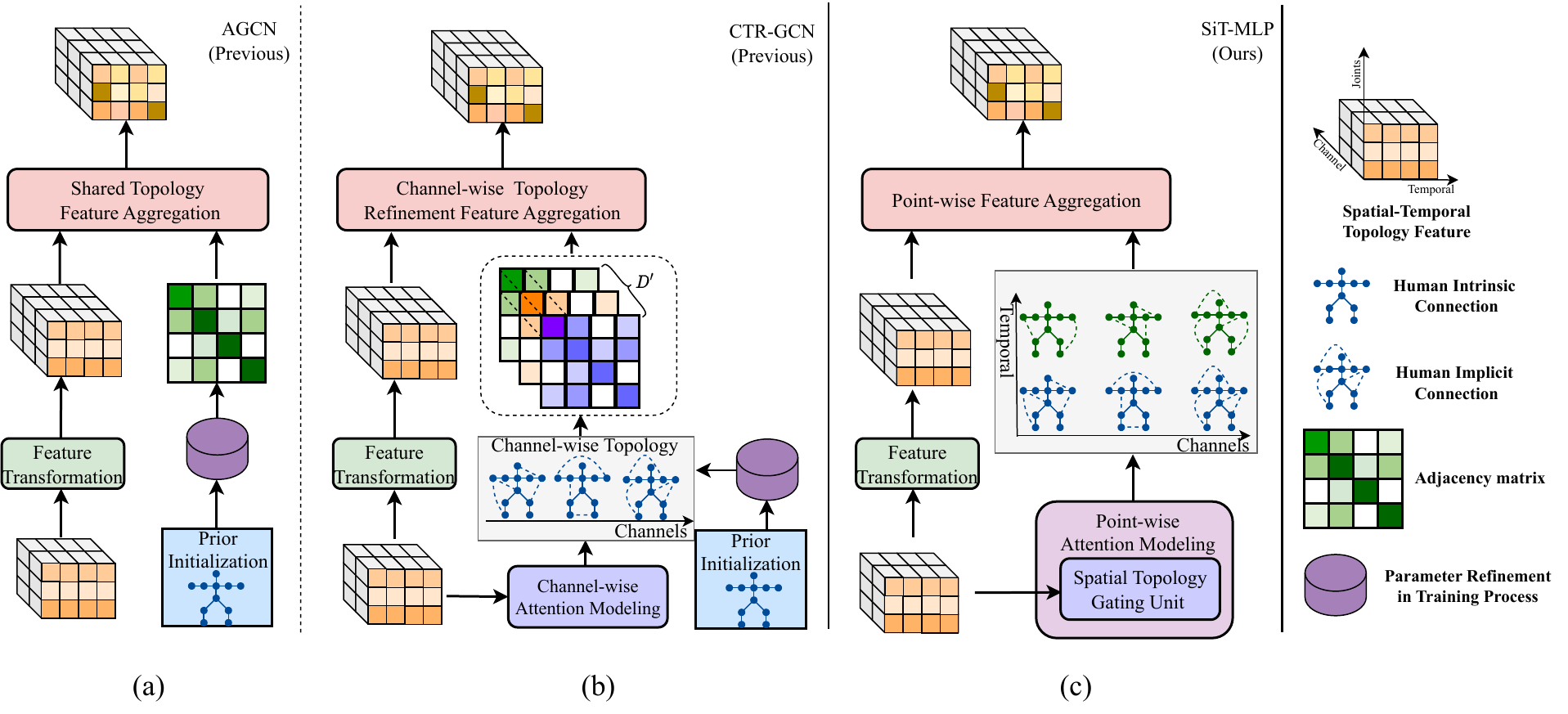}
    \caption{The spatial modeling structure of different approaches: (a) the normally sample-generic modeling module; (b) the channel-wise topology refinement modeling module; (c) the proposed Spatial Topology Gating Unit.}
    \label{fig: C}
\end{figure*}

Moreover, another reason we believe the SOTA GCN-based approaches have been able to achieve such success is the inclusion of sample-specific attention modeling based on feature interaction mechanisms. As demonstrated in Fig. \ref{fig: C} (a), in some previous SOTA methods \cite{yan2018spatial,shi2019two}, the feature transformation module and the topology modeling module are independent. The learned topology attention is sample-generic. The static networks and shared topology connections limit the capability of the model. To break these limitations and learn sample-specific topology dependency, some feature interaction mechanism was adopted in other works \cite{chen2021channel,liu2020disentangling}. The interaction between features can allow the network to adjust the output according to the input samples dynamically. As shown in Fig. \ref{fig: C} (b), \cite{chen2021channel} model sample-specific attention according to the input features. The attention is denoted as affinity matrices, and this sample-specific topology attention is non-shared in the channel dimension. The adaptation of this channel-wise attention improves the performance of the network significantly.

Although the sample-specific modeling obtained good performance for action recognition, the temporal pooling operations adopted in \cite{chen2021channel} result in the channel-wise attention shared in the temporal dimension. Human action includes coupled spatial-temporal relationships. Therefore, the modeling of temporal-wise topology attention is necessary. For example, during the process of drinking coffee, the implicit connections between the head joints and the right-hand joints will gradually become stronger. The temporal-wise attention can model topological relationships over time dynamically and channel-wise attention can enhance the distinctiveness of features. The combination of these two kinds of attention can be denoted as point-wise attention, which is non-shared in the channel and temporal dimensions. \textbf{The modeling of point-wise sample-specific attention can encourage the network to fully explore the spatial-temporal relations of the input action}. The gating unit\cite{liu2021pay} can achieve such modeling without complex aggregations and huge parameters.

With the above findings, in this work, we present a simple yet effective topology feature learning model, SiT-MLP, the first model using MLP to address skeleton-based action recognition. As shown in Fig. \ref{fig: C} (c), in SiT-MLP, to model sample-specific spatial correlations, we propose the Spatial Topology Gating Unit(STGU), which is an MLP-baed structure without any priors and introduces a new gate-based feature interaction mechanism. On the one hand, in STGU, features can be activated point-to-point by the generated attention map. The point-wise sample-specific topologies provide independent relationships between any two joints in the temporal and channel dimensions. On the other hand, compared to the self-attention mechanism \cite{vaswani2017attention} containing up to 3rd-order interactions (e,g., $\textit{q}_i\textit{k}_j\textit{v}_k$), the gate mechanism containing up to 2nd-order interactions (e.g., $\textit{z}_i\textit{z}_j$ ), reduces the computational consumption significantly.

We summarize our main contributions as follows:
\begin{itemize}
    \item We show an insight that skeleton-based human action recognition can be modeled simply without elaborate priors. This prior-free characteristic can make the corresponding algorithm generalize easier, avoiding the complex design of the prior in different scenarios.  
    \item We propose the Spatial Topology Gating Unit (STGU) to learn the point-wise sample-specific topology features with simple aggregation. 
    \item Based on STGU, we propose SiT-MLP to tackle skeleton-based action recognition, which falls into the category of MLP-based methods. To the best of our knowledge, SiT-MLP is the first MLP-based model in skeleton-based action recognition.
    \item Extensive experiments on three large-scale datasets demonstrate the proposed SiT-MLP achieves favorable results against the previous schemes. Moreover, SiT-MLP also reduces the parameters and computing resources significantly.
\end{itemize}
\section{Related Work}
\subsection{Skeleton-based Action Recognition}
In early studies, Recurrent neural network (RNN) based methods \cite{du2015hierarchical, song2017end, zhang2017view} and Convolution neural network (CNN) based methods \cite{li2018co, liu2017enhanced, dang2020dwnet,banerjee2020fuzzy} are popular choices for solving the skeleton-based action recognition problem.  However, the methods mentioned above overlook human topology and the spatial interactions between joints. 

\textbf{GCN-based Methods.} To effectively handle the skeleton sequence, GCN, which has a strong ability to extract features from non-Euclidean data, has become a more popular option in this field. ST-GCN \cite{yan2018spatial} first applies GCN to model the joint correlations and significantly boosts the performance of skeleton-based action recognition. However, the inherent topology structure limits the GCN in capturing long-range relations. To break this limitation,  MS-G3D \cite{liu2020disentangling} and STIGCN \cite{huang2020spatio} adopt multi-scale graph topologies to GCNs to model multi-range joint relations. To further aggregate the global spatial features, most subsequent works \cite{shi2019two, shi2020decoupled, zhang2020semantics, ye2020dynamic} have adopted a learnable topology for action recognition.

 Considering that the spatial relations that different actions depend on may be different, many works \cite{zhang2020semantics, chen2021channel, shi2019two, ye2020dynamic, zhou2023learning, liu2023skeleton} tend to construct sample-specific topology connections. SGN \cite{zhang2020semantics} enhances topology learning with the self-attention mechanism, which models correlation between two joints given corresponding features. CTR-GCN \cite{chen2021channel}   learns different topologies dynamically and aggregates joint features in different channels effectively for skeleton-based action recognition. LKA-GCN \cite{liu2023skeleton} employs a skeleton large kernel attention operator, which can enlarge the receptive field and improve channel adaptability. DD-GCN \cite{li2023dd} constructs the directed diffusion graph for action modeling and introduces the activity partition strategy. FR-Head \cite{zhou2023learning} introduces contrastive learning to enhance the sample-specific features learning. CD-JBF-GCN \cite{tu2022joint} designs a novel correlation-driven joint-bone fusion graph convolutional network to fully explore the latent sample-specific correlation between joints and bones. HD-GCN \cite{lee2023hierarchically} effectively constructs an HD-Graph by decomposing every joint node into several sets to extract major adjacent and distant edges.

Recently, some hypergraph-based methods \cite{hao2021hypergraph, zhu2022selective} have also achieved competitive results. Hyper-GNN \cite{hao2021hypergraph} adopts a hypergraph to capture both spatial-temporal information and high-order dependencies for skeleton-based action recognition. SHGCN \cite{zhu2022selective} adaptively chooses the optimal hypergraph structure to represent skeleton actions instead of using a fixed graph structure or predefined hypergraph structure.

\textbf{Transformer-based Methods.} Attempts have been made recently to solve this problem using Transformers. They mainly focused on handling the challenge brought by the extra temporal dimension. ST-TR \cite{plizzari2021skeleton} adopts a two-stream model consisting of spatial and temporal self-attention for modeling intra- and inter-frame correlations, respectively. GAT \cite{zhang2022graph} adapted transformer and graph-aware masks to model the spatial relationship between any joint. DSTA\cite{shi2020decoupled} employed a Transformer to model the spatial and temporal dimensions alternately. TranSkeleton \cite{liu2023transkeleton} proposes a novel partition-aggregation temporal transformer, which works with hierarchical temporal partition and aggregation, and can capture both long-range dependencies and subtle temporal structures effectively. However, there is still a performance gap between GCN-based methods and Transformer-based methods. The reasons for this gap may be the two intrinsic factors: (1) transformer-based methods maintain the sequence length throughout the model, which leads to huge redundancy as the input sequences are generally long. The redundant information carried by the attention matrix, which increases with the length of the sequence, prevents the network from focusing on the key to the current action. (2) The absence of sample-generic modeling makes it different to capture the common features of input sequences, especially when trained on limited-scale skeleton datasets. 

In summary, both GCN-based methods and Transformer-based methods improve the performance by contrasting complex aggregation mechanisms, leading to the heavyweight of the networks.

\subsection{Vision MLP}
Inspired by the successful adaptation of the Transformer \cite{vaswani2017attention} in the computer vision field \cite{dosovitskiy2020image, liu2021swin}, many works have researched the intrinsic mechanism of self-attention. On the one hand, some works \cite{tolstikhin2021mlp, touvron2022resmlp, hou2022vision, tang2022sparse, zhang2021morphmlp} explore the feasibility of MLP for the global modeling of the sequences. Mixer-MLP \cite{tolstikhin2021mlp} and ResMLP \cite{touvron2022resmlp} adopted an MLP architecture instead of the self-attention mechanism to achieve the cross-token communication. ViP \cite{hou2022vision} encodes the feature representations along the height and width dimensions with linear projections. SMLP \cite{tang2022sparse} adds a feature extraction along channel dimension and a depthwise convolution in front of each block to enhance the local modeling to ViP. MorphMLP \cite{zhang2021morphmlp} achieves competitive performance with recent SOTA methods, demonstrating that the MLP-like backbone is also suitable for video recognition.

On the other hand, some other studies believe that the dynamic weights brought by the attention map provide a larger parameter space for the network. Thus, to break the limitation of the static parameters, the attention mechanism is introduced to the MLP-based network. GMLP \cite{liu2021pay} adopts the gating mechanism to the MLP-based architecture to improve the generalization of the network. GSwin \cite{go2023gswin} combines parameter efficiency and performance with locality and hierarchy in image recognition. VAN \cite{guo2022visual} indicates that the key characteristic of attention methods is adaptively adjusting output based on the input feature but not the normalized attention map.

Due to the absence of the adaptation of the skeleton data and the modeling of the complex spatiotemporal relations, these classical MLP-based approaches in computer vision can’t obtain discriminative representations. Therefore, combining MLP and previous methods in the task of action recognition, we propose the SiT-MLP to tackle the skeleton-based action recognition without priors and complex aggregations.

\section{Methods} 
In this section, we first introduce the architecture of our SiT-MLP. Then we elaborate on our Spatial Topology Gating Unit (STGU), which is the main contribution of our SiT-MLP. Finally, we compare our STGU with other GCN-based methods and analyze why our method works.

\subsection{Model Architecture}
\begin{figure}
    \centering
    \includegraphics[scale=0.57]{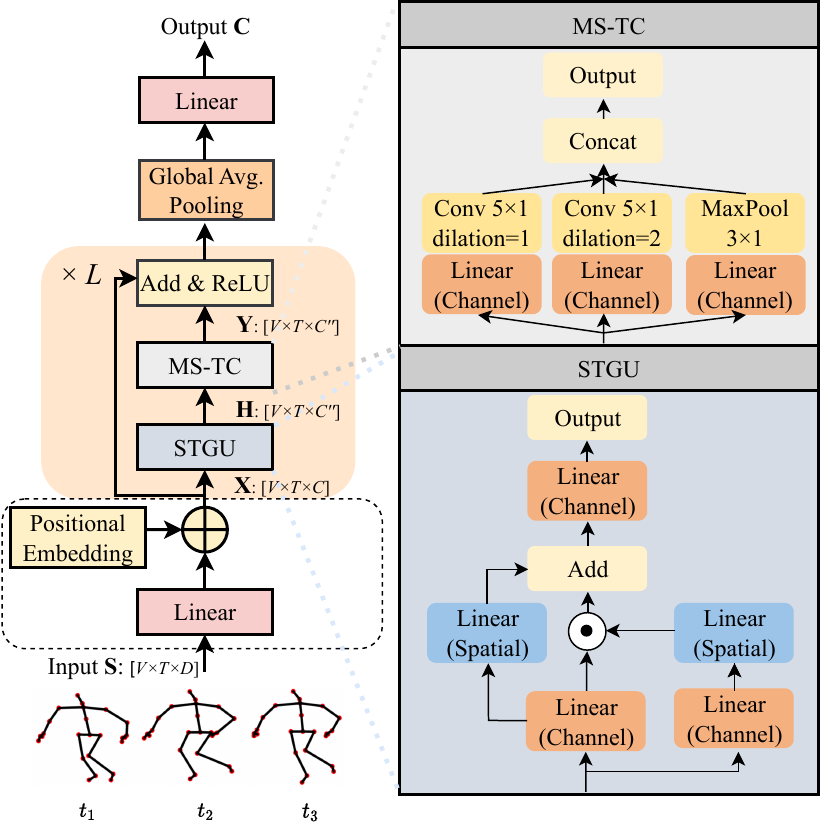}
    \caption{Model architecture overview and illustration. The embedding block is adopted to retain the positional information. The STGU module captures the spatial dependency, and the MS-TC module aggregates the temporal information. The global average pooling layer is used to aggregate the global spatial-temporal joint information for the final linear classifier.}
    \label{fig: M}
\end{figure}

We construct a novel but simple network SiT-MLP for skeleton-based human action recognition. As shown in Fig. \ref{fig: M}, the entire network consists of an embedding block and five basic blocks, followed by a global average pooling and a linear classifier to predict action categories. It is more difficult for joint spatial-temporal optimization \cite{zhang2021morphmlp}. Therefore, following previous works \cite{chen2021channel,yan2018spatial,shi2019two}, we place the spatial STGU and temporal MS-TC module in the sequential style in the basic block. Particularly, SiT-MLP is built by stacking several of these basic blocks. The global average pooling layer is used to aggregate the spatial-temporal information for the final linear classifier. Although there are some convolutions in the temporal modeling module, our SiT-MLP is mainly composed of liner layers, falling into the category of MLP-based method.

\textbf{Embedding Block.} Our STGU is an MLP-based structure, and MLPs are insensitive to the permutation of positions. To break the permutation equivariance in our SiT-MLP, an embedding block is used to retain positional information. 

As shown in Fig. \ref{fig: M}, the input sequence $\boldsymbol{S}$ is processed by a learnable embedding, which linearly projects body joint $s_t \in \mathbb{R}^{V \times D}$ through a linear layer to the hidden dimension $C$. To identify the positional information of the joints, we add pose embeddings (\textit{PE}) to the projected features. The learnable parameters \textit{PE} are shared across times. We refer to the output of the learnable linear projection $\boldsymbol{X}^{(0)}=\left \{x^{(0)}_1,...,x^{(0)}_t,...,x^{(0)}_T\right \} \in \mathbb{R}^{V\times T \times C}$ as:
\begin{equation}
x^{(0)}_t=s_t\textbf{W} + \textit{PE}, 
\end{equation}
where $x^{(0)}_t ,\textit{PE} \in \mathbb{R}^{V \times C}; \textbf{W} \in \mathbb{R}^{D \times C}$; \textit{t} is the time index. Moreover, $\boldsymbol{X}^{(0)}$ is also the input of the first stacked basic block.
 
\textbf{Spatial Modeling.} As shown in Fig \ref{fig: M}, in a spatial modeling module, we just use an STGU block to extract correlations between human joints. The details of STGU are demonstrated in section \ref{STGU}. To improve the diversity of feature space and avoid the risk of overfitting, we follow the previous approaches \cite{go2023gswin} adopting the multi-head mechanism in \cite{vaswani2017attention} in our STGU. The multi-head mechanism enables the network to compute simultaneously without additional consumption. Note that we omit the multi-head operation following for simplicity. Different from previous approaches \cite{chen2021channel, yan2018spatial, shi2019two, liu2020disentangling} that required three parallel branches to introduce different human priors, we only adopt one branch for the spatial information extraction without priors. As the above-mentioned decomposition, SiT-MLP can reduce parameters and computational overhead by one-third in the process of spatial modeling.

\begin{figure*} [htbp]
     \centering
    \includegraphics[scale=0.66]{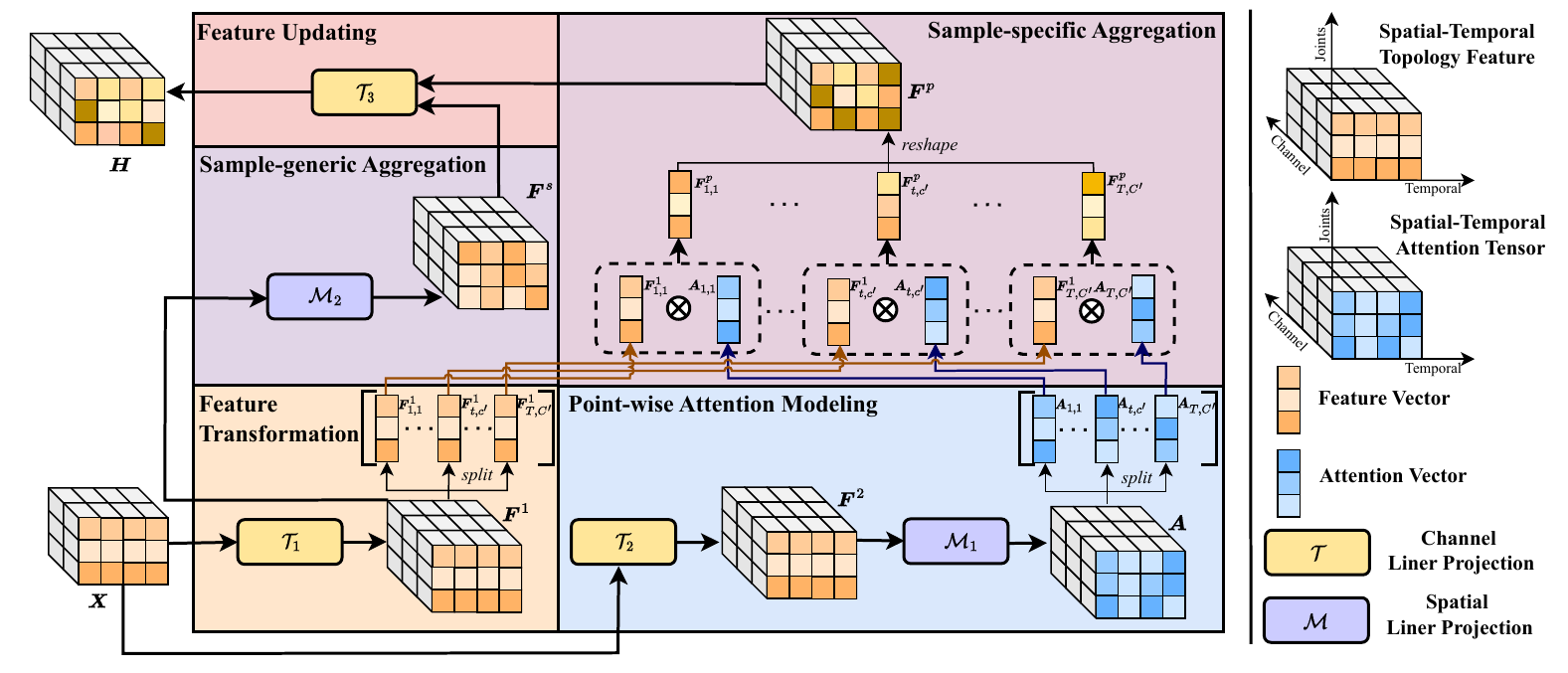}
    \caption{Framework of the proposed spatial topology gating unit. Feature transformation aims at transforming input features into latent high-dimensional feature space. Point-wise attention modeling builds the entire independent topology attention. Sample-specific aggregation aims to select dynamic features for the current sample. Sample-generic aggregation is for capturing the common feature between all samples. Feature updating aims at fusing and updating the feature after aggregation.}
    \label{fig: F}
\end{figure*}

\textbf{Temporl Modeling.} As shown in the grey block in Fig. \ref{fig: M}, we adopt the Multi-Scale Temporal Convolution (MS-TC) \cite{liu2020disentangling} to aggregate the temporal dependency of the human pose. This module contains three parallel branches with a combination of different kernel sizes and dilation rates, and max-pooling to capture multi-scale temporal information. The extracted features of different branches are concatenated. The MS-TC can capture action characteristics over varying time lengths, which allows the model to understand both short-term rapid movements and long-term, slowly evolving actions.

SiT-MLP is constructed by stacking STGU and Temporal Convolution layers alternately as follows:
\begin{align}
     \boldsymbol{H}^{(l)} & = \text{STGU}(\boldsymbol{X}^{(l)}), \\
     \boldsymbol{Y}^{(l)} & = \text{MS-TC}(\boldsymbol{H}^{(l)}), \\
     \boldsymbol{X}^{(l+1)} & = \text{ReLU}(\boldsymbol{Y}^{(l)} + \boldsymbol{X}^{(l)}).
\end{align}
where $l$ is the index of the stacked basic block. As shown in Fig. \ref{fig: M}, for training stability, the standard residual connections are used in each basic block. Moreover, to add non-linearity, a ReLU activation layer is adopted after each basic block after spatial and temporal modeling modules. To manage the temporal dynamics inherent in action recognition, the temporal dimension is halved at the 2-nd and 4-th blocks by strided temporal convolution for sequential frame feature fusion.

\subsection{Spatial Topology Gating Unit}
\label{STGU}

The general framework of our STGU is shown in Fig. \ref{fig: F}. Specifically, Our STGU contains five parts; (1) Feature transformation aims at transforming input features into latent high-dimensional feature space; (2) Point-wise attention modeling builds the entire independent topology attention; (3) Sample-specific aggregation aims to select dynamic features for the current sample; (4) Sample-generic aggregation is for capturing the common feature between all samples. (5) The feature updating aims at fusing and updating the feature after aggregation. 

\begin{figure}
    \centering
    \includegraphics[scale=0.56]{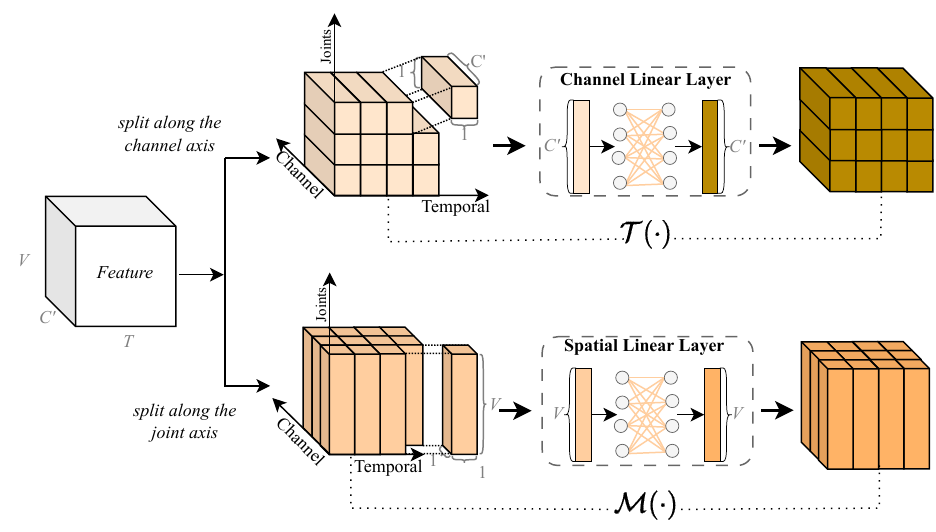}
    \caption{The details of the projection $\mathcal{T}(.)$ and $\mathcal{M}(.)$ in our STGU. Here, for ease of understanding, we define the input feature $\in \mathbb{R}^{V \times T \times C'}$}
    \label{fig: liner_layer}
\end{figure}

Here, we introduce the details of the projection $\mathcal{T}(\cdot)$ and $\mathcal{M}(\cdot)$ in our STGU, which are shown in Fig. \ref{fig: liner_layer}. Channel linear layer $\mathcal{T}(\cdot)$ acts on channels of feature, allowing the aggregation of the spatial features between different channels and operating on each joint and frame independently. Spatial linear layer $\mathcal{M}(\cdot)$ acts on joints of feature, capturing the spatial across all joints and operating on each channel and frame independently. Next, we will describe each block in detail. 

\textbf{Feature Transformation.} As in previous methods, the input spatial-temporal feature of our STGU is $\boldsymbol{X} \in \mathbb{R}^{V \times T \times{C}}$. Feature transformation aims at transforming input features into latent high-dimensional feature space via $\mathcal{T}_1(\cdot)$. We adopt a simple linear transformation here as a linear layer, which can be formulated as 
\begin{equation}
    {\boldsymbol{F}}^{1}=\mathcal{T}_1(\boldsymbol{X})=\sigma(\boldsymbol{X}\mathbf{W}_1),
\end{equation}
where $\boldsymbol{F}^{1} \in \mathbb{R}^{V \times T \times C'}$ is the transformed feature and $\mathbf{W}_1 \in \mathbb{R}^{C \times C'}$ weight matrix of the linear channel projection. The details of transformation $\mathcal{T}(\cdot)$ are shown in Fig. \ref{fig: liner_layer}. The channel liner layer $\mathcal{M}(\cdot)$ can project the features and treat each joint in each frame individually. The linear channel projection can be represented as a $1 \times 1$ convolution layer. And $\sigma$ is an activation function such as GeLU \cite{hendrycks2016gaussian}. 

\textbf{Point-wise Attention Modeling.} As shown in the blue block in Fig. \ref{fig: F}, 
we aim at constructing dynamic parameters, which can provide point-wise attention for the transformed feature. First, we adopt the same operation in feature transformation to project the input to the latent feature space. Then we use a spatial linear layer to encourage communication of each joint. This cross-joint communication can provide a global receptive field of the current pose for the attention map. The attention modeling process can be formulated as 
\begin{equation}
\label{eq: attention}
    \boldsymbol{A} = \mathcal{M}_1(\mathcal{T}_2(\boldsymbol{X}))=\mathbf{W}_A(\sigma(\boldsymbol{X}\mathbf{W}_2)),
\end{equation}
Here, $\boldsymbol{A} \in \mathbb{R}^{V \times T \times C'}$ is the attention map, which is dynamically parameterized based on the input feature. The details of the spatial linear layer $\mathcal{M}(\cdot)$ are shown in Fig. \ref{fig: liner_layer}, and $\mathbf{W}_A \in \mathbb{R}^{V\times V}$ is the learnable parameters of the spatial linear layer. There is not any dimension pooling in this process, so our attention map still retains the diversity of features in the channel dimension and the seriality in the temporal dimension. In this way, the point in the attention $\boldsymbol{A}$ can carry the global spatial-temporal information of all the input sequences. 

\textbf{Sample-specific Aggregation.} The transformed feature $\boldsymbol{F}^{1}$ and defined attention map $\boldsymbol{A}$ can be split across the channel dimension and the temporal dimension. In other words, both the spatial-temporal topology feature and spatial-temporal attention can be regarded as a set of vectors, which can be formulated as follows
\begin{equation}
    \begin{split}
    \boldsymbol{F}^{1} = \{\boldsymbol{F}^{1}_{1,1}, \, \cdot \cdot \cdot, \, \boldsymbol{F}^{1}_{t,c'}, \, \cdot \cdot \cdot, \, \boldsymbol{F}^{1}_{T,C'}\}  ,\\
    \boldsymbol{A} = \{\boldsymbol{A}_{1,1}, \, \cdot \cdot \cdot, \, \boldsymbol{A}_{t,c'}, \, \cdot \cdot \cdot, \, \boldsymbol{A}_{T,C'}\} ,\\
    \end{split}
\end{equation}
Here, $t$ and $c'$ are the indexes of features in the temporal dimension and channel dimension respectively. $\boldsymbol{F}^{1}_{t,c'}, \in \mathbb{R}^{V}$ is the feature vector, and $\boldsymbol{A}_{t,c'} \in \mathbb{R}^{V}$ is the attention vector. Since reshaping was not performed in the previous modeling process, $\boldsymbol{F}$ and $\boldsymbol{A}$ are perfectly aligned in both the channel dimension and temporal dimension. Each topology vector is able to find the corresponding attention vector. Then we can adopt the attention vector to aggregate the feature vector point-to-point.
\begin{equation}
    \boldsymbol{F}^{p}_{t,c'}=\boldsymbol{F}^{1}_{t,c'} \odot \boldsymbol{A}_{t,c'},
\end{equation}
where, $\boldsymbol{F}^{p}_{t,c'} \in \mathbb{R}^{V}$ is the topology feature after aggregation. Since the attention $\boldsymbol{A}_{t,c'}$ is generated from the input feature, which will change dynamically depending on different inputs. This gating unit, which is denoted as element-wise multiplication, allows the network to select features of interest based on the generated attention map. In this way, our STGU can adaptively select the discriminative features and ignore noisy responses automatically. Then we reshape the set of feature vectors after sample-specific aggregation into the topology feature.
\begin{equation}
    \{\boldsymbol{F}^{p}_{1,1}, \, \cdot \cdot \cdot, \, \boldsymbol{F}^{p}_{t,c'}, \, \cdot \cdot \cdot, \, \boldsymbol{F}^{p}_{T,C'}\} = \boldsymbol{F}^{p}, 
\end{equation}
Here, $\boldsymbol{F}^{p} \in \mathbb{R}^{V \times T \times C'}$ is the feature after sample-specific aggregation. This aggregation is capable of capturing complex spatial interactions across joints. In  $\boldsymbol{F}^{p}$, each vector $\boldsymbol{F}^{p}_{t,c'}$ is obtained by activating $\boldsymbol{F}^{1}_{t,c'}$ and $\boldsymbol{A}_{t,c'}$ at the corresponding positions.  And these corresponding positions denote the same temporal and channel indexes. Thus, our sample-specific aggregation is point-wise aggregation. In this way, we successfully model the temporal-wise attention dynamically while retaining the channel-wise topology attention. Moreover, the entire point-wise aggregation can be computed in parallel without much computational overhead.

\textbf{Sample-generic Aggregation.} To encourage our STGU to model the connectivity between joints and learn the common features of all samples, we use another spatial liner layer to perform shared-topology feature modeling. 
\begin{equation}
\label{eq: shared}
    \boldsymbol{F}^{s} = \mathcal{M}_2(\boldsymbol{F}^{1}) = \mathbf{W}_S\boldsymbol{F}^{1},
\end{equation}
where $\boldsymbol{F}^{s} \in \mathbb{R}^{V \times T \times C'}$ is the feature after sample-generic aggregation. Although the generation process of $\boldsymbol{F}^{s}$ and $\boldsymbol{A}$ is similar, the modeling is different. We adopt the attention $\boldsymbol{A}$ to select the discriminative features through the gating unit. And $\boldsymbol{F}^{s}$ without any special definitions and feature interactions denotes the feature after sample-generic aggregation, which models the common dependency across all samples. 

\textbf{Feature Updating.} Feature updating aims at fusing $\boldsymbol{F}^{p}$ after sample-specific aggregation and $\boldsymbol{F}^{s}$ after sample-generic aggregation and updating the feature. The total modeling process can be formulated as:
\begin{equation}
\label{eq: y}
    \boldsymbol{H} = \mathcal{T}_3(\boldsymbol{F}^{p}, \boldsymbol{F}^{s})=\sigma((\boldsymbol{F}^{p} + \boldsymbol{F}^{s})\mathbf{W}_O)
\end{equation}
Here, $\mathbf{W}_O \in \mathbb{R}^{C' \times C''}$ is the weight of the channel projection layer we adopt to update the feature. And $\boldsymbol{H} \in \mathbb{R}^{V \times T \times C''}$ is the output of our STGU.

The static sample-generic modeling pays more attention to modeling human connectivity, and the point-wise attention balances the joint-level and implicit connectivity modeling well. By fusing the features after two aggregations, the output $\boldsymbol{H}$ of our STGU updates the information of the input $\boldsymbol{X}$ successfully without complex aggregations.

 
The pseudo-code of our STFU is shown in Algorithm \ref{alg: 1}. For training stability, we initialize $\mathbf{W}_A$ in equation \ref{eq: attention} as near-zero values, meaning that $\boldsymbol{A} \approx \boldsymbol{F}^{p} \approx \mathbf{0}$ at the beginning of training. In this way, the equation \ref{eq: y} can be written as 
\begin{equation}
    \boldsymbol{H} = \sigma(\boldsymbol{F}^{s}\mathbf{W}_O) = \sigma(\mathbf{W}_S\boldsymbol{F}^{1}\mathbf{W}_O),
\end{equation}

This initialization ensures each STGU block behaves like a regular GCN at the early stage of training. $\mathbf{W}_S$ is the aggregation weight matrix and $\mathbf{W}_O$ is the updating weight matrix. We find that such initialization plays an important role in model convergence. With constant training and parameter updating, the STGU gradually injects specific spatial information related to the current sample.

\begin{algorithm}[t]
    \caption{Pseudo-code for the STGU (Pytorch-like)}
    \label{alg: 1}
    \begin{lstlisting}[language=Python, basicstyle=\ttfamily, breaklines=true,keywordstyle=\bfseries\color{NavyBlue},morekeywords={},emph={self}, emphstyle=\bfseries\color{Rhodamine},commentstyle=\itshape\color{black!50!white},stringstyle=\bfseries\color{PineGreen!90!black},columns=flexible,numbers=left,numbersep=2em,numberstyle=\footnotesize]  
# x: input tensor of shape (B, T, V, C)
# d_model: out-channel of our STGU
# num_joints: number of the joints

def STGU(x, d_model, num_joints):
  shortcut = x
  x = norm(x, axis="channel")
  x = proj(x, 2*d_model, axis="channel")
  F_1, F_2 = split(x, axis="channel")
  A = proj(F_2, num_joints, axis="spatial", init_weight = 0)
  F_p = F_1 * A
  F_s = proj(F_1, num_joints, axis="spatial")
  H = proj((F_s + F_p), d_model, axis="channel")
  return shortcut + H
    \end{lstlisting}
\end{algorithm}

\subsection{Comparison of modeling processes with GCN-based approaches}
The most significant difference between SiT-MLP and GCN-based approaches in the modeling processes is the spatial modeling module. In previous methods \cite{yan2018spatial, cheng2020skeleton, chen2021channel, chi2022infogcn, plizzari2021skeleton}, whether the adjacency matrices in the GCN blocks or the normalized affinity matrix in the self-attention \cite{si2019attention}, it is an intuitive representation of human connections. Each element in the normalized 2D matrix can denote the intensity of this connection. The connection may be explicit or implicit. The network can aggregate the information of related joints in the affinity matrix. 

\begin{figure}[t]
    \centering
    \includegraphics[scale=0.58]{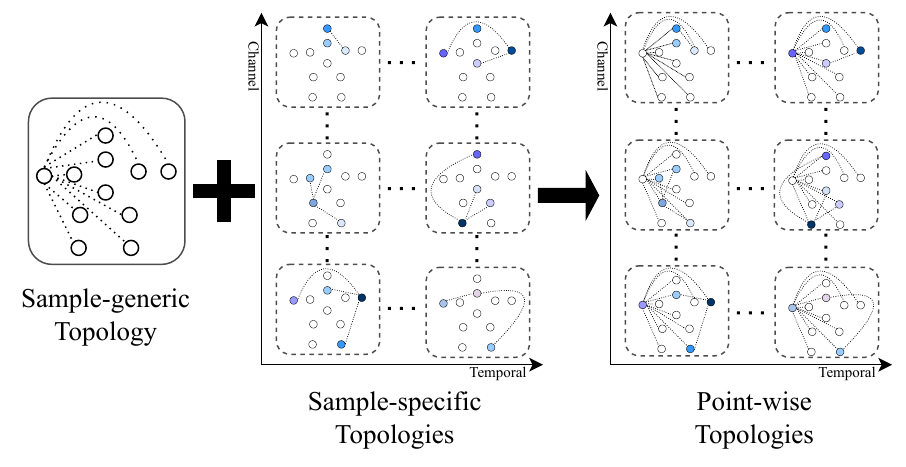}
    \caption{The topology modeling process of our STGU. The dotted line represents implicit connections. The darker color of the joints indicates more attention to the joints.}
    \label{fig: P}
\end{figure}

However, as shown in Fig. \ref{fig: P}, in the sample-specific aggregation module, each element in our attention map indicates the importance of the corresponding joints. This attention shows diversity in the channel dimension and adaptability in the temporal dimension. Moreover, since the attention is modeled through a spatial liner layer $\mathcal{M}(\cdot)$, there is a spatial communication between joints. Every element can carry information about other joints, which can be represented as an implicit connection. Thus, the sample-specific aggregation can be viewed as a joint representation of the importance of joints and implicit connections.

There are some similarities between our method and previous methods \cite{yan2018spatial, shi2019two}. The sample-generic aggregation module in STGU plays the same role as conventional GCNs. The spatial liner layer can model the relations between any two joints and the weight can be denoted as a joint-to-joint affinity matrix. The proposed SiT-MLP can aggregate the global spatial dependency through the spatial liner layer. The sample-generic aggregation models the common connections across all samples.

Finally, after point-wise feature aggregation and global connectivity feature aggregation, the topology features learned in our SiT-MLP contain the spatial-temporal relations of the input sample.



\begin{table*}[ht]
    \centering
    \caption{Action classification performance comparison against SOTA methods on the NTU RGB+D 60 and NTU RGB+D 120 dataset. Infogcn \cite{chi2022infogcn} adopted additional MMD losses for better recognition performance. * indicates that we retrain the models using their officially released code.}
    \resizebox{\textwidth}{!}
    {
    \begin{tabular}{@{}l|lcccccc@{}}
     \toprule
    \multirow{2}*{Type} & \multirow{2}*{Methods} & \multirow{2}*{Parameters (M)} &\multirow{2}*{FLOPs (G)} &\multicolumn{2}{c}{NTU RGB+D 60} &  \multicolumn{2}{c}{NTU RGB+D 120} \\
         &   &   &  & X-Sub(\%) & X-View(\%)    &  X-Sub(\%)  & X-Set(\%)   \\\hline
    \multirow{2}*{RNN} & VA-LSTM \cite{zhang2017view} & -  & - & 79.4 & 87.6 & - & - \\
    & AGC-LSTM \cite{si2019attention} & 22.9 ($\downarrow$ 97.3\%) & - & 89.2 & 95.0 & - & - \\\hline
    \multirow{4}*{CNN} & DWNet \cite{dang2020dwnet} & - & - & 84.1 & 89.8 & - & - \\
    & HCN \cite{li2018co} & - & - & 86.5 & 91.1 & - & - \\
    & VA-CNN \cite{zhang2019view} & 24.1 ($\downarrow$ 97.5\%) & - & 88.7 & 94.3 & - & - \\
    & Ta-CNN+ \cite{xu2022topology} & 1.1 ($\downarrow$ 45.5\%) & - & 90.7 & 95.1 & 85.7 & 87.3 \\\hline
    \multirow{7}*{GCN} & ST-GCN \cite{yan2018spatial} & 3.1 ($\downarrow$ 80.6\%) & - & 81.5 & 88.3 & 70.7 & 73.2 \\
    & Shift-GCN \cite{cheng2020skeleton} & 0.7 ($\downarrow$ 14.3\%) & 2.5 & 90.7 & 96.5 & 85.9 & 87.6 \\
    & Graph2Net \cite{wu2021graph2net} & 0.9 ($\downarrow$ 22.2\%) & - & 90.1 & 96.0 & 86.0 & 87.6 \\
    & DC-GCN+ADG \cite{cheng2020decoupling} &2.5 ($\downarrow$ 96.0\%) & 2.8 & 90.8 & 96.6 & 86.5 & 88.1 \\
    & MS-G3D \cite{liu2020disentangling} & 2.8 ($\downarrow$ 78.5\%) & 5.2 & 91.5 & 96.2 & 86.9 & 88.4 \\
    & Hyper-GNN \cite{hao2021hypergraph} & - & - &89.5 &95.7 & - & - \\
    & MST-GCN \cite{chen2021multi} & 2.8 ($\downarrow$ 78.5\%) & {-} & 91.5 & 96.6 & 87.5 & 88.8 \\\hline 
    \multirow{11}*{GCN+Att} &SGN \cite{zhang2020semantics} & 0.7 ($\downarrow$14.3\%) & 0.8 & 89.0 & 94.5 & 79.2 & 81.5 \\
    & 2s-AGCN \cite{shi2019two} & 3.5 ($\downarrow$ 82.9\%) & 3.9 & 88.5 & 95.1 & 82.9 & 84.9 \\
    & CD-JBF-GCN ( \cite{tu2022joint} & - & - & 89.0 & 95.4 & - & -\\
    & LKA-GCN \cite{liu2023skeleton} & - & - & 90.7 & 96.1 & 86.3 & 87.8\\
    & Dynamic GCN \cite{ye2020dynamic} & 14.4 ($\downarrow$ 95.8\%) & - & 91.5 & 96.0 & 87.3 & 88.6 \\
    & EfficientGCN-B4 \cite{song2022constructing} & 2.0 ($\downarrow$ 70.0\%) & 2.0 & 91.7 & 95.7 & 88.3 & 89.1 \\
    & CTR-GCN* \cite{chen2021channel} & 1.4 ($\downarrow$ 60.0\%)& 1.8  & 92.4 & 96.8 & 88.9 & {90.5}\\
    & InfoGCN*  \cite{chi2022infogcn} & 1.5 ($\downarrow$ 62.5\%) & 1.7 & {92.3}  & {96.7} & 89.2 & 90.7  \\ 
    & DD-GCN \cite{li2023dd} & - & - & 92.6 &96.9 & 88.9 & 90.2\\
    & FR-Head \cite{zhou2023learning} & 2.0 ($\downarrow$ 70.0\%) & - & 92.8 & 96.8 & 89.5 & 90.9 \\ 
    & HD-GCN \cite{lee2023hierarchically} & 1.7 ($\downarrow$ 64.7\%) & 1.6 & \textbf{93.0} & \textbf{97.2} & \textbf{89.8} & \textbf{91.2} \\
    \hline
    \multirow{4}*{Transformer} & GAT \cite{zhang2022graph}  & 5.9 ($\downarrow$
    89.8\%) & - & 89.0 & 95.2 & 84.0 & 86.1 \\
    & ST-TR \cite{plizzari2021skeleton} & 12.1 ($\downarrow$ 95.0\%) & 259.4 & 89.9 & 96.1 & 82.7 & 84.7 \\
    & DSTA \cite{shi2020decoupled} & 4.1 ($\downarrow$ 85.3\%) & 64.7 & 91.5 & 96.4 & 86.6 & 89.0 \\
    & TranSkeleton \cite{liu2023transkeleton} & 2.2 ($\downarrow$ 72.7\%)  & 9.2 & 92.8 & 97.0 & 89.4 & 90.5 \\\hline
    \multirow{2}*{MLP} & SiT-MLP (Joint Only) & \textbf{0.6} & 0.7 &90.0 & 95.0 & 83.9 & 85.7 \\
    & SiT-MLP (4 ensemble) & \textbf{0.6} & \textbf{0.7} &{92.3} & 96.8 & {89.0} & {90.5}  \\
    \bottomrule
    \end{tabular}
    }
    \label{tab:ntu_result_label}
\end{table*}
\section{Experiment}
To demonstrate the effectiveness of our proposed SiT-MLP, we conduct skeleton-based action recognition on three large-scale datasets. We compare our model with powerful baselines on both performance and complexity. Moreover,  we conduct ablation studies in order to examine the effect of individual components. To further evaluate the SiT-MLP, a more extensive failure case analysis is provided. In the end, some visualization will be shown to evidence our motivation and contribution.

\subsection{Datasets}

\textbf{NTU RGB+D 60.} NTU RGB+D 60 \cite{shahroudy2016ntu} is a large-scale dataset for skeleton-based human action recognition. It contains 56880 videos performed by 40 volunteers. The action sequences can be categorized into 60 classes. Following the authors of this dataset recommendation, we process this dataset into two benchmarks: cross-subject(X-sub) and cross-view(X-view). In the cross-subject setting, sequences of 20 subjects are for training, and the sequences of the rest 20 subjects are for validation. In the cross-view setting, skeleton sequences are split by camera views. Samples from two camera views are used for training, and the rest are used for evaluation.

\textbf{NTU RGB+D 120.} NTU RGB+D 120 \cite{liu2019ntu} dataset adds 57367 new skeleton sequences and 60 new action classes to the original NTU RGB+D 60 dataset. There are 32 various video configurations in it, each of which depicts a different location and background. The authors offered the cross-subject(X-sub) and cross-setup(X-set) as two benchmark evaluations. In the cross-subject setting, sequences from 53 subjects are for training, and sequences from the other 53 subjects are for testing. In the cross-setup setting, skeleton sequences are split by setup ID. Samples from even set-up IDs are used for training, and the odd setup IDs are used for evaluation.

\textbf{Northwestern-UCLA.} Northwestern-UCLA dataset \cite{wang2014cross} contains 1494 video clips of 10 different actions captured from three Kinect cameras. We follow the same evaluation protocol in \cite{wang2014cross}: the first two cameras for training and the other for testing.

\subsection{Implementation Details}
Following the previous experimental setting \cite{chen2021channel}, we train our model with SGD with momentum 0.9, weight decay 0.0004, and training batch size 64. The total training epoch is set to 90, and a warmup strategy is used in the first five 5 epochs to make the training process more stable. We set the learning rate to decay with cosine annealing \cite{loshchilov2016sgdr}, with a base learning rate of 0.1 and an end learning rate of 0.0001. The standard Cross-Entropy loss is adopted to optimize our model. For NTU RGB+D 60 and NTU RGB+D 120 datasets, we adopt the data preprocessing method from \cite{chen2021channel}. For the Northwestern-UCLA dataset, the batch size is 16, and we process the data followed by \cite{cheng2020skeleton}. Our project is based on Pytorch \cite{paszke2019pytorch}, and the training and testing experiments are conducted on four NVIDIA GTX 1080Ti GPUs.

\subsection{Comparison With the State-of-the-Art}
For a fair comparison of network performance, we follow the most recent SOTA approaches\cite{chen2021channel, cheng2020decoupling} to train models on 4 skeleton modalities(\textbf{joint}, \textbf{bone}, \textbf{joint motion}, \textbf{bone motion}) and report the recognition performance of the ensemble. As shown in the TABLE \ref{tab:ntu_result_label}, the comparisons include the RNN/CNN-based methods \cite{zhang2017view, si2019attention,dang2020dwnet,li2018co,zhang2019view,xu2022topology}, GCN-based methods \cite{yan2018spatial,cheng2020skeleton,cheng2020decoupling,liu2020disentangling,hao2021hypergraph,chen2021multi,zhang2020semantics,ye2020dynamic,shi2019two,chen2021channel,song2022constructing,chi2022infogcn,wu2021graph2net},Transformer-based methods \cite{zhang2022graph,plizzari2021skeleton,shi2020decoupled}.

We compare our results with previous SOTA approaches on NTU RGB+D 60 in TABLE \ref{tab:ntu_result_label}. The X-sub protocol accuracy and X-view protocol accuracy of our SiT-MLP are 92.3\% and 96.8\% respectively. The comparison of NTU RGB+D 120 is also shown in the TABLE \ref{tab:ntu_result_label}. The X-sub protocol accuracy and X-set protocol accuracy of our SiT-MLP are 89.0\% and 90.5\% respectively, which are also favorable results. The comparison of the Northwestern-UCLA dataset is shown in TABLE \ref{tab:ucla_result_label}, the accuracy on the validation set is 96.5\%. SiT-MLP achieves competitive results, which surpass most of the existing methods. For instance, we largely outperform the lightweight SGN \cite{zhang2020semantics} model by 9.8\% and 9.0\% on NTU120 X-sub and X-view benchmark respectively. Although the performance of our SiT-MLP falls behind the recent state-of-the-art GCN-based methods to some extent, HD-GCN \cite{lee2023hierarchically} effectively constructs an HD-Graph by decomposing every joint node into several sets to model high-order relations for better performance. FR-Head \cite{zhou2023learning} introduces contrastive learning to enhance sample-specific feature learning. These methods all sacrifice computational efficiency in exchange for higher accuracy. Without complex aggregations and elaborate priors, the proposed SiT-MLP significantly reduces parameters and consumption with competitive performance on various benchmarks.

Furthermore, we compare the throughput and memory usage of the SiT-MLP and previous methods. Throughput indicates the number of input sequences that can be processed per second, and memory usage indicates the resource required in the inference phase. As shown in Table \ref{tab: throughput}, the proposed SiT-MLP can process the skeleton sequences more efficiently with less memory usage. For instance, our SiT-MLP can process 395.5 more sequences per second with competitive performance. Therefore, SiT-MLP is more promising in terms of real-time processing and its deployment in resource-constrained environments.

All the experimental results proved that our SiT-MLP can extract spatial-temporal co-occurrence features more effectively.

\begin{figure*}
    \centering
    \includegraphics[scale=0.33]{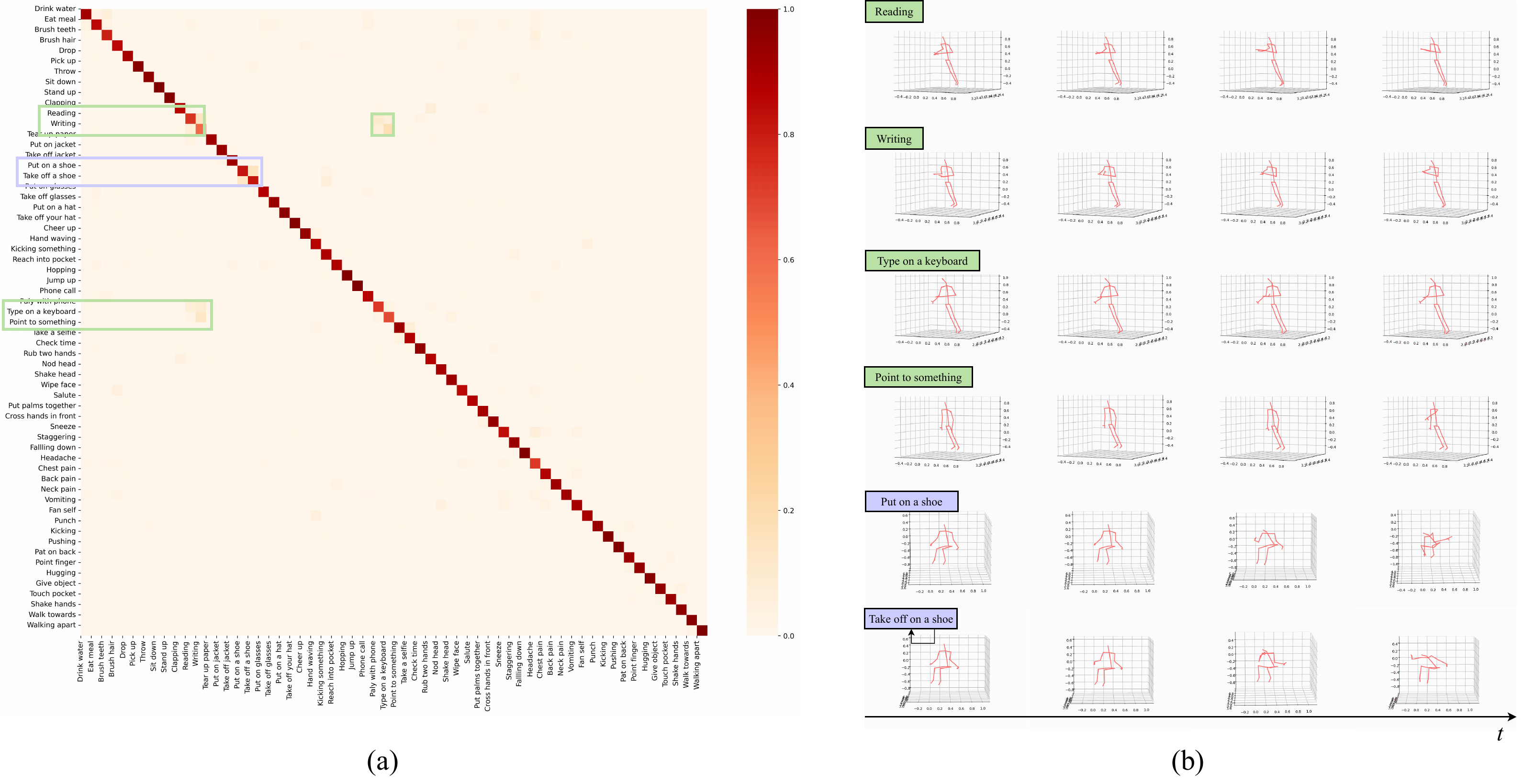}
    \caption{(a) The confusion matrix on the X-sub benchmark of NTU RGB+D 60. The vertical coordinate is the true label, and the horizontal is the prediction. (b) The visualization of 3D skeletons of the failure case.}
    \label{fig: confusion}
\end{figure*}

\begin{table}[h]
\caption{Classification performance on the Northwestern-UCLA dataset.}
  \centering
  \resizebox{0.4\textwidth}{!}{
  \begin{tabular}{@{}l|lccc@{}}
    \toprule
   Type & Methods  &  UCLA Acc(\%) \\
    \midrule
        \multirow{2}*{RNN} & TS-LSTM \cite{lee2017ensemble} & 89.2 \\
        & 2s-AGC-LSTM \cite{si2019attention} &93.3 \\\hline
        \multirow{2}*{CNN} & VA-CNN \cite{zhang2019view}  & 90.7\\
        & Ta-CNN \cite{xu2022topology} & 96.1 \\\hline
        \multirow{6}*{GCN} & Shift-GCN & 94.5 \\
        & Graph2Net \cite{wu2021graph2net} & 95.3 \\
        & CTR-GCN \cite{chen2021channel} & {96.5} \\
        & InfoGCN w/ MMD losses \cite{chi2022infogcn} & \textbf{97.0} \\
        & DD-GCN \cite{li2023dd} & 96.7 \\
        & FR-Head \cite{zhou2023learning} & 96.8 \\
        & HD-GCN \cite{lee2023hierarchically} & 96.9 \\
        \hline
        \multirow{2}*{MLP} & SiT-MLP (Joint Only) & 95.7 \\
        & SiT-MLP (4 ensemble) & {96.5} \\
     \bottomrule
  \end{tabular}
  }
  \label{tab:ucla_result_label}
\end{table}

\begin{table}[]
    \centering
    \caption{The comparisons with throughput and memory usage between the proposed SiT-MLP with previous methods.}
    \begin{tabular}{c|cc}
        \toprule
         Method & Throughput (sequence/s) $\uparrow$ & Memory Usage (G) $\downarrow$ \\
         \midrule
         2S-AGCN \cite{shi2019two} & 397.7 & 1.9 \\
         MS-G3D \cite{liu2020disentangling} & 249.2 & 1.9 \\
         CTR-GCN \cite{chen2021channel} & 296.1 & 1.8 \\
         InfoGCN \cite{chi2022infogcn} & 307.4 & 1.9 \\
         HD-GCN \cite{lee2023hierarchically} & 97.5 & 1.8\\
         \hline
         SiT-MLP (ours) & 457.0 & 1.6 \\
         \bottomrule
    \end{tabular}
    \label{tab: throughput}
\end{table}

\subsection{Ablation Studies}

To examine the effect of individual components of SiT-MLP, we compare the classification accuracy of different configurations of our model. All experimental ablation studies are conducted on NTU RGD+D 60 dataset cross-subject benchmarks with joint modal information. 

\begin{table}[]
    \centering
    \caption{Constructing SiT-MLP from the vanilla baseline.}
    \begin{tabular}{@{}lccc@{}}
        \toprule
        Model &Parameters  &Acc(\%) \\
        \midrule
        MH-STGU Only & 0.5M  & 76.9\\
        MS-TC Only & 0.3M & 86.4 \\
        SiT-MLP (STGU + MS-TC) & 0.6M & \textbf{90.0} \\
        \bottomrule
    \end{tabular}
    \label{tab: components}
\end{table}

\textbf{The design of SiT-MLP} As shown in Fig. \ref{fig: M}, the basic block of our SiT-MLP is the combination of STGU and MS-TC. To verify that our SiT-MLP can achieve spatial-temporal optimization and extract the spatial-temporal co-occurrence feature, we explore the ablation with only the STGU module and the MS-TC module. As shown in TABLE \ref{tab: components}, the temporal modeling module MS-TC improves accuracy by 10.1 \%, and our STGU contributes to the final performance, with a significant improvement of 3.6\% accuracy. Our SiT-MLP, the combination of STGU and MS-TC, can 
capture the spatial-temporal co-occurrence features well.

\begin{table}[]
    \centering
        \caption{The effectiveness of the MH-STGU components.}
    \begin{tabular}{@{}lccc@{}}
        \toprule
        Method  &Acc(\%) \\
        \midrule
         STGU &\textbf{90.0}  \\
         STGU w/o Sample-generic Aggregation &88.5 \\
         STGU w/o Sample-specific Topology Aggregation &89.0 \\
         STGU w/o Temporal-wise Modeling &89.8\\
         STGU w/o Channel-wise Modeling &89.3 \\\hline
         STGU w Prior Initialization & \textbf{90.0}\\
         \bottomrule
    \end{tabular}
    \label{tab: ablation}
\end{table}

\textbf{The effectiveness of the STGU components.} The innovation of our SiT-MLP mainly focuses on the STGU module. In this section, we verify the effectiveness of the STGU in individual components. The experimental details are shown in TABLE \ref{tab: ablation}.

\textit{(1) Evaluation of the Sample-generic Aggregation Module:} This module is for modeling common features across samples. If we prune the module, the network just focuses on sample-specific modeling and ignores sample-generic modeling. Fully dynamic networks lack modeling of common features, and we can observe that the recognition accuracy of the network drops by 1.5\%. It shows that the shared topology modeling can capture rich common feature information across all samples.

\textit{(2) Evaluation of the Sample-specific Aggregation Module:} Contrary to what was said before, if we prune the dynamic modeling module, the whole network is statically parameterized, which can't adjust the parameter of the network based on the input sample. From TABLE \ref{tab: components}, we can find that compared with the original STGU, the performance has dropped approximately by 1\%. It demonstrates that sample-specific modeling and sample-generic modeling play an important role in recognition performance.
 
\textit{(3) Evaluation of Temporal-wise Modeling:} To further verify the effectiveness of our point-wise attention modeling We follow the recent approach \cite{chen2021channel} and adopt the temporal pooling operation in attention modeling. In this way, we force the network to model temporal-shared topologies. From TABLE \ref{tab: components}, the performance has decreased by 0.2\%. Thus, the temporal-wise topology modeling is working for skeleton-based action recognition.

\textit{(4) Evaluation of Channel-wise Modeling:} Similar to the above-mentioned, we adopt the channel pooling operation in attention modeling, which forces the network to model channel-shared topology. In this way, the performance has decreased by 0.7\%. The channel-shared topology reduces the distinctiveness of features. Thus, modeling of channel-wise topology is vital for learning distinctive topology.

From a combination of results \textit{(3)} and \textit{(4)}, either temporal-wise or channel-wise topology modeling has an impact on the performance of the network. Therefore, the point-wise sample-specific topology modeling, which not only achieves adaptability in the channel dimension but also adaptability in the temporal dimension, achieves great performance.

\textit{(5) Evaluation of Prior Initialization:} To verify our SiT-MLP is insensitive to the initialization, we initialize the spatial linear layer in equation \ref{eq: shared} with the human body's natural connections. We initialize the weight of the linear layer with the binary outward connection matrix. As shown in TABLE \ref{tab: components}, compared with our SiT-MLP, whose spatial mixing linear layers are initialized as diagonal matrices, the classification accuracy of SiT-MLP with prior initialization is flat. The results confirm that our SiT-MLP can learn spatial information without any well-designed initialization.

\begin{table}[]
    \centering
    \caption{The comparisons between the 3D skeletons captured from three Microsoft Kinect v2 cameras and skeletons extracted from the RGB videos of joint modality.}
    \begin{tabular}{c|cc}
        \toprule
        Method  & NTU120/Xsub (Kinect) & NTU120/Xsub (RGB) \\
       \midrule
        ST-GCN \cite{yan2018spatial} & 82.1 & 80.1 ($\downarrow$ 2.0) \\
        2S-AGCN \cite{shi2019two} & 82.8 & 80.2 ($\downarrow$ 2.6) \\
        CTR-GCN \cite{shi2019two} & 84.0 & 82.2 ($\downarrow$ 1.8) \\
        \hline
        SiT-MLP (ours) & 83.9 & 83.2 ($\downarrow$ 0.7) \\
        \bottomrule
    \end{tabular}
    \label{tab: rgb}
\end{table}

\textbf{Evaluation of the Generalization.} To quantitatively analyze the model’s robustness and generalization capabilities, we measure the performance of SiT-MLP in real-world scenarios. We extract human skeletons from the RGB-based videos, where actions occur in complex environments with background noise. We use Faster-RCNN \cite{ren2015faster} for human detection and HRNet \cite{sun2019deep} for human pose estimation. As shown in TABLE \ref{tab: rgb}, the performance of SiT-MLP is decreased by 0.7\%. Since there is no need to introduce priors of the human body connections to the network, the proposed SiT-MLP demonstrates greater generality. Compared with the existing GCN-based method, the accuracy drop of our SiT-MLP is relatively small, which proves that our method is more general and robust.

\subsection{Discussion}

\begin{figure}
    \centering
    \includegraphics[scale=0.61]{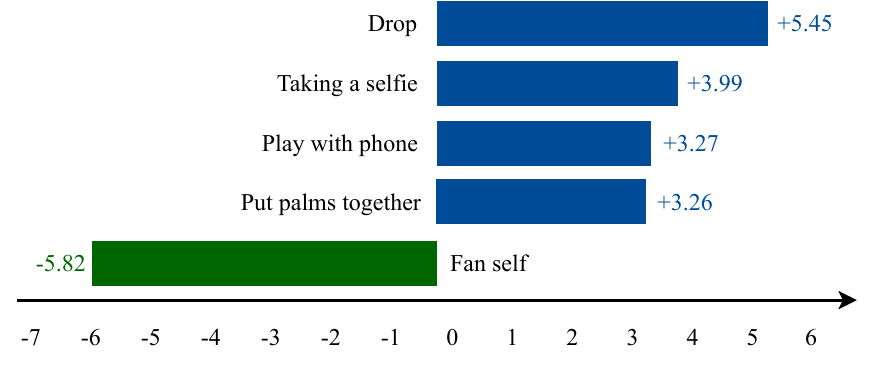}
    \caption{Action classes with accuracy differences higher than 3\% between CTR-GCN and our method on NTU60 X-sub benchmark.}
    \label{fig: Experiment}
\end{figure}

To further analyze the results of our SiT-MLP, we visualize the confusion matrix on the X-sub benchmark of NTU RGB+D 60 in Fig. \ref{fig: confusion}(a). The main confusion in our approach lies in two parts: actions of "\textit{Reading}", "\textit{Writing}", "\textit{Type on a keyboard}", "\textit{Point to something}", and actions of "\textit{Put on a shoe}", "\textit{Take off a shoe}". There are two main reasons for our confusion. First, as shown in Fig. \ref{fig: confusion}(b), due to the absence of information omitted in skeletonization (hand skeleton, head, related objects, etc.), some actions have been difficult to classify. Second, our SiT-MLP is designed to pay attention to capturing global relations. Because of the absence of local modeling of the human body, SiT-MLP failed to classify the fine-grained actions well. The main difference in actions of "\textit{Reading}" and "\textit{Writing}" may focus on the details of the hand skeleton. This lack of modeling of fine-grained local information leads our SiT-MLP to confuse these two actions.

Moreover, to further analyze the failure cases, we compare the classification results of SiT-MLP (90.0) with CTR-GCN (89.6), which is a powerful GCN-based method.  As shown in Fig. \ref{fig: Experiment}, there is a decrease in the accuracy of actions "\textit{Fan self}". We argue the reason for this decrease is that the proposed SiT-MLP can’t tell the repetitive actions well, like "\textit{Fan self}". The SiT-MLP can model the frame-wise attention. However, because of the repetition within the action, the frame-wise attention modeling makes it hard to capture the general feature of the periodicity of actions, leading to a performance drop.  Conversely, thanks to point-wise topology feature learning, the proposed SiT-MLP can adjust the spatial relations between joints dynamically over time. Actions "\textit{Type on a keyboard}", "\textit{Drop}", "\textit{Taking a selfie}", "\textit{Play with phone}", and "\textit{Put palms together}" are prolonged actions, which are not repetitive and contain more complex spatial relations. The reason for the increase in these actions is that frame-wise attention provides independent spatial relations in the spatial dimension.

\subsection{Qualitative Analysis} 
In this section, we make some visualizations and analyses to support our work.

\begin{figure*}
    \centering
    \includegraphics[scale=0.129]{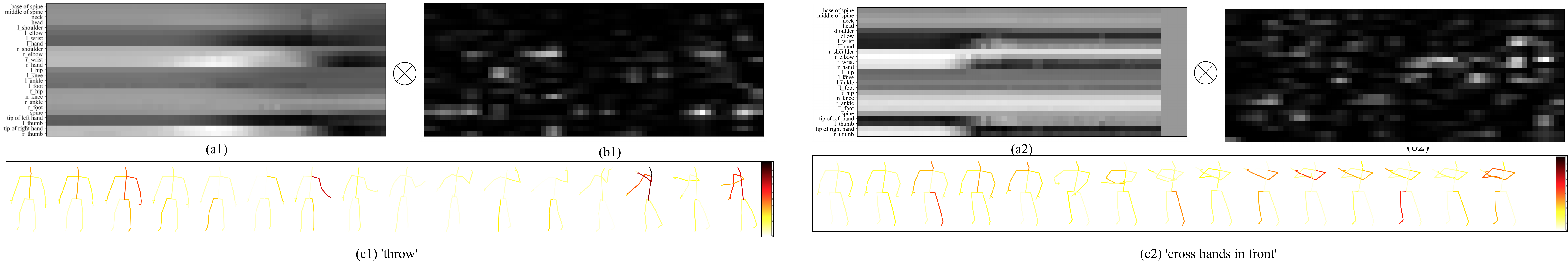}
    \caption{Heatmap visualization of the proposed SiT-MLP: (a) motion time series images, (b) Grad-CAM images, and (c) skeleton motion highlighted with color for recognition reasons. The darker the color, the more attention the area receives.}
    \label{fig: cam}
\end{figure*}

\textbf{Grad-CAM.} As shown in Fig. \ref{fig: cam}, we select two classical actions ‘throw’ and ‘cross hands in front’, and visualize both the original images in Fig. \ref{fig: cam}(a) and corresponding Grad-CAM \cite{selvaraju2017grad} images in Fig. \ref{fig: cam}(b). To show the ROI of the SiT-MLP more directly, we visualize the results of the Grad-CAM on the input skeleton sequence in Fig. \ref{fig: cam}(c). It can be observed that the SiT-MLP always relies on the key area of the input sequence to predict the categories. For the action of ‘throw’, the activation areas are more focused on the movements of upper body joints in the keyframe where the action occurs. Furthermore, for the action of ‘cross hand in front’, the activation areas are focused on the joints of the hands and arms, which can be viewed as a more semantically informative region.

\begin{figure}
    \centering
    \includegraphics[scale=0.265]{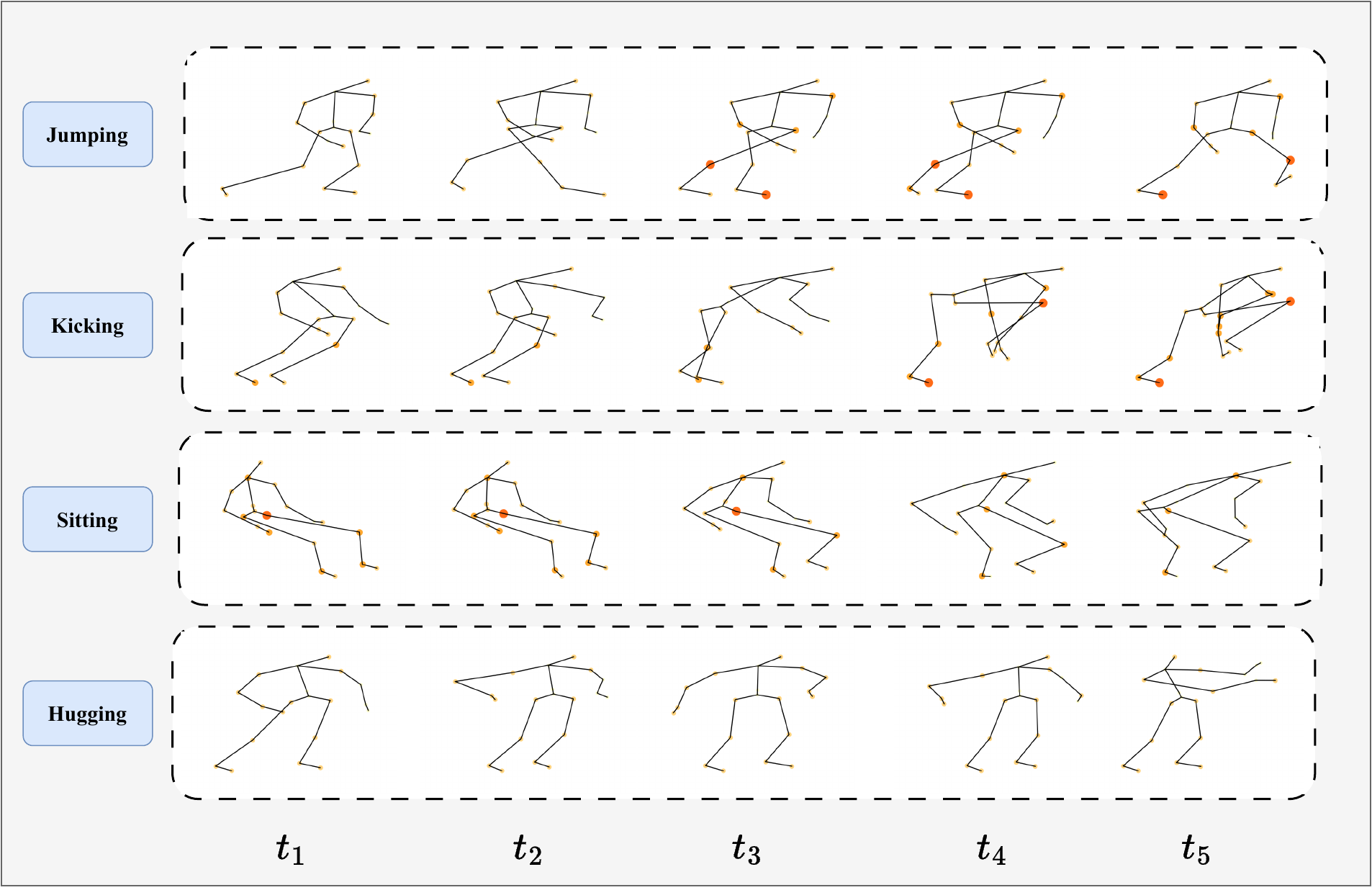}
    \caption{Examples of sample-specific topology attention modeling of STGU. Lines indicate the skeletal connections between human bodies. The darker color and larger size of the joints indicate that the more attention it receives.}
    \label{fig: T}
\end{figure}

\textbf{Sample-specific topology learning.} We illustrate the sample-specific topology attention of four typical action samples in Fig \ref{fig: T}. The darker color and larger size of the joints indicate that the more attention it receives. During the jumping, it is intuitive that our SiT-MLP pays more attention to the tendency of the legs and knees to move. Furthermore, in an action like kicking, it is kinematically correct to focus on the knee and the opposite foot at the same time. For actions like sitting, our model pays more attention to the hip. This attention slowly diminishes as the human begins to rise. During the hugging, Our SiT-MLP can also focus equally on the trajectories of the limbs. Above all, our SiT-MLP realizes not only spatial but also temporal topology awareness. 

\section{Limitations}
Although the proposed SiT-MLP achieves inspiring performance on various benchmarks, however, according to TABLE \ref{tab:ntu_result_label} and \ref{tab:ucla_result_label}, SiT-MLP falls behind HD-GCN. This gap may be caused by poor performance in the recognition of fine-grained and repetitive actions. Specifically, our SiT-MLP is designed to capture global spatial relations but reduce the ability to model the local information in the body part, leading to a performance drop for fine-grained action recognition. In this case, the possible solution is to introduce the complementary modeling of local and global information to assist network learning. In addition, because frame-wise attention modeling hinders the capture of the general feature of the periodicity of action to some extent, the SiT-MLP can't recognize repetitive action well. It is necessary to design between frame-wise topology attention to recognize periodic actions. In further work, we will continue to study and solve these two drawbacks.

\section{Conclusion}
In this paper, we try to use MLPs to address skeleton-based action recognition. We propose STGU, the first MLP-based structure for spatial dependency modeling without extra priors and complex aggregations. In STGU, a new feature interaction mechanism, the gate mechanism, is introduced to this task, which can implement sample-specific and point-wise attention modeling. Without the elaborate priors, our SiT-MLP shows great generalizability, which can be generalized to other fields of action recognition. Without complex aggregations, the proposed SiT-MLP significantly reduces parameters with competitive performance on various benchmarks. Moreover, the SiT-MLP is more promising in terms of real-time processing and its deployment in resource-constrained environments. We hope the proposed SiT-MLP will help the community with some inspiration for the task of skeleton-based action recognition.
\section{Acknowledgement}
This work was supported partly by the National Natural Science Foundation of China (Grant No. 62173045), partly by the Natural Science Foundation of Hainan Province (Grant No. 622RC675) and the Fundamental Research Funds for the Central Universities (Grant No. 2020XD-A04-3). 

\bibliographystyle{IEEEtran}
\bibliography{ref}


\end{document}